\documentclass{article}

\usepackage{graphicx}
\usepackage{amsmath,amssymb}
\DeclareMathOperator*{\argmax}{arg\,max}
\DeclareMathOperator{\EX}{\mathbb{E}}%
\usepackage{graphicx}
\usepackage{caption}
\usepackage{subcaption}
\usepackage{wrapfig}

\usepackage[final]{neurips_2021}

\usepackage[utf8]{inputenc} 
\usepackage[T1]{fontenc}    
\usepackage{hyperref}       
\usepackage{url}            
\usepackage{booktabs}       
\usepackage{amsfonts}       
\usepackage{nicefrac}       
\usepackage{microtype}      
\usepackage{xcolor} 
\usepackage{authblk}

\title{Offline Contextual Bandits for Wireless \\ Network Optimization}
\author[1,2,*]{Miguel Suau}
\author[1]{Alexandros Agapitos}
\author[1]{David Lynch}
\author[1]{Derek Farrell}
\author[1]{\\Mingqi Zhou}
\author[1]{Aleksandar Milenovic}
\affil[1]{Huawei Ireland Research Center}
\affil[2]{Delft University of Technology} \affil[*]{\texttt{m.suaudecastro@tudelft.nl}}
\begin{document}

\maketitle

\begin{abstract}
The explosion in mobile data traffic together with the ever-increasing expectations for higher quality of service call for the development of AI algorithms for wireless network optimization. In this paper, we investigate how to learn policies that can automatically adjust the configuration parameters of every cell in the network in response to the changes in the user demand. Our solution combines existent methods for offline learning and adapts them in a principled way to overcome crucial challenges arising in this context. Empirical results suggest that our proposed method will achieve important performance gains when deployed in the real network while satisfying practical constrains on computational efficiency.
\end{abstract}

\section{Introduction}
The evolution of \emph{wireless networks} from 2G, 3G, 4G and now to 5G, in response to users’ insatiable appetite for greater capacity and lower latency, has been driven by continuous innovations in technology and infrastructure. New data hungry services now available to the user (such as online gaming, video streaming and virtual reality) are causing an explosion in mobile data traffic. Furthermore, the usage growth translates into increasingly complex wireless networks. These factors render network operation and maintenance infeasible. Currently, network operators 
adjust the configuration parameters (CPs) of every cell in the network through a trial and error process. Manually adjusting CPs is time consuming process and typically results in highly suboptimal settings. Long lead times cause previously performant CPs to become stale relative to dynamic usage demand and meteorological conditions. 
Moreover, the scale of the network, with up to 10K inter-dependent cells and many CPs per cell makes ad hoc solutions impractical. 


Here we investigate a data-driven approach to wireless network optimization. We learn policies that select the appropriate CPs for each cell, given information characterizing the current network state. Our end goal is to optimize the average \emph{throughput} (TP) per user using only a \emph{static dataset} of past interactions \citep{Metevier2019Offline, sachdeva2020off,levine2020offline} with the real network. The key challenges addressed in this paper are summarized below:
\begin{enumerate}
    \item \textbf{Large action space:} Our goal is to control 14 CPs in total. Any combination of CP values within the allowed range is a candidate solution. Optimizing TP given only a static data set is thus not trivial since the resulting action space is very large and the CPs may have complex, possibly non-linear, effects on TP.
    \item \textbf{Data sparsity and covariate shift:} 
    Due to the large combinatorial action space, it is impractical to obtain a densely sampled dataset. Certain parts of the state-action space will be inevitably under-represented in the logged data. Moreover, given that we want to improve over the logging policy, the action distribution induced by the learned policy should be different from the logged distribution. This phenomenon, known as covariate shift \citep{bickel2009discriminative, quinonero2009dataset}, only exacerbates the problem of sample sparsity since the actions we are interested in will likely be poorly supported in the data.
    \item \textbf{Computational efficiency:} 
    The number of cells in the network is large (2945) and updated CPs must be pushed to all cells simultaneously every hour. As such, due to hardware constraints, it is essential that our solution is computationally efficient.
\end{enumerate}

The main contributions of this paper are: 
\begin{enumerate}
    \item We formulate of the wireless network optimization problem as an \textbf{offline contextual bandit} \citep{Auer2002TheNM,Langford2008Epoch}.
    \item We propose a \textbf{hybrid method} that combines a \textbf{policy network} learned via importance sampling to suggest preliminary actions, followed by \textbf{gradient-based local fine-tuning} on a differentiable model of the TP.
    \item We introduce a \textbf{modular neural network architecture} for TP prediction that ensures a balanced representation of all input modalities and \textbf{enhances the sensitivity} of the model to the CPs.
    \item Finally, to guarantee the reliability of the method against model bias, we perform \textbf{counterfactual data augmentation} to help our model generalize over unseen actions, and apply a \textbf{penalty} to the objective function to prevent the optimization algorithm from finding solutions for which the TP predictions are \textbf{uncertain}.
\end{enumerate}
The rest of the paper is organized as follows. Section \ref{sec:preliminaries} introduces the problem formulation and mathematical notation. We also provide a short description of the dataset used to train and evaluate our policies. Section \ref{sec:existing_approaches} reviews the two most popular offline learning approaches for the contextual bandit setting. In section \ref{sec:method}, we present our method and describe how we addressed the above challenges. Finally, the results of our experiments are reported in Section \ref{sec:evaluation}, from which key conclusions are drawn.

\section{Preliminaries}\label{sec:preliminaries}

\subsection{Problem definition}

Let us call $s \in \mathcal{S}$ the \emph{network state} (sometimes referred to as \emph{context} in the bandits literature), which is specified by performance management (PM) counters, engineering parameters (EP) and other contextual information describing the network environment. $\mathcal{A}$ is the action space of 14 dimensions, one for each of the CPs to be optimized,
such that each of the infinitely many actions $a \in \mathcal{A} = \mathcal{A}_1 \times \ldots \mathcal{A}_{14}$ represents a particular parameter configuration at a specific time.  The task consists in finding the policy $\pi(s,a)=P(a|s)$ that maximizes a certain objective function,
\begin{equation}
\small
    J(\pi) = \EX_{s \sim \mathcal{S}, a \sim \pi(s, a)} \left[R(s,a) = r \right]
\end{equation}
and $r$ is the feedback provided by the system. In our case, $R(s,a)$ is an unknown function that models the TP per user given the action $a$ and the network state $s$. As opposed to the more general reinforcement learning framework, the bandit formulation assumes that actions have no temporal dependencies. While this might introduce approximation errors, it also simplifies the optimization problem. Moreover, this simplification is motivated by the fact that CP changes occur only once per day in our static dataset, which might dilute temporal dependencies from hour-to-hour. 

\subsection{Dataset Description}\label{sec:dataset}

As mentioned in the introduction, we are given a static dataset of past system interactions collected using a fixed policy $\pi_0$, known as the logging policy,
\begin{equation}
\small
    \mathcal{D}_{\pi_0} = \{s_i, a_i, r_i\}_{i=1}^N.
\end{equation}
The dataset was retrieved from a real 4G network. The network contained 2945 cells served by 1030 radio base stations.
The dataset covered a period of 9 days during which changes to 14 CPs were made \emph{daily} to each cell between midnight and 2 AM. The 14 CPs were chosen because they have the largest impact on TP. A total of 295K CP changes were registered. The effect of these changes on the network was monitored by hourly performance management (PM) counters. Other information regarding the engineering parameters (EP) of the antennas, their physical location, and the time of the day was also included in the dataset.

\section{Existing Approaches}\label{sec:existing_approaches}

In the offline learning setting, we are not allowed to explore and assess actions on the real network as is normally assumed for standard contextual bandits. This poses a fundamental challenge for the learning algorithm. On the one hand, we want the new policy to outperform the logging policy, which requires it to take different actions from those observed in the dataset. On the other hand, the outcomes for these hypothetical actions are not present in the dataset. All we can do is learn from $\mathcal{D}$ those actions that lead to high-payoffs in throughput with the expectation that our policy will be able to generalize effectively over unseen contexts. See \citet{levine2020offline} for an extensive discussion on this matter.
Below we describe the most popular solutions for the offline learning problem in the contextual bandit setting. 

\subsection{Direct Method}\label{sec:directmethod}
The \emph{direct method} \citep{Beygelzimer2009Offset} consists of learning a regression model that for every state $s$ can predict the reward for taking any of the actions available, $\hat{R}(s,a) \approx R(s,a)$ for all $a \in \mathcal{A}$. This can be then used to compute a policy $\pi(s)$ as 
\begin{equation}
\small
    \pi(s) = \argmax_{a \in \mathcal{A}} \hat{R}(s,a). 
    \label{eq:dm}
\end{equation}
It is clear that the performance of DM depends very much on the accuracy of the reward model. Unfortunately, since this is only trained on $\mathcal{D}$, its estimates might be biased from the true reward values when the number of samples is too small. Note that a biased model is particularly problematic in this case, because the \emph{argmax} in equation \eqref{eq:dm} forces the policy to select the actions for which the reward estimates are the largest, which may or may not correspond to those actions that maximize the true reward. 
\subsection{Inverse Propensity Score}
An alternative to the DM is to obtain an estimate of the objective function $J(\pi)$ by applying importance sampling \citep{horvitz1952generalization} on experiences drawn from $\mathcal{D}$,
\begin{equation}
\small
    J(\pi_\theta) 
    \approx \frac{1}{N} \sum_{(x_i,a_i,r_i) \in \mathcal{D}} \frac{\pi(s_i,a_i)}{\pi_0(s_i,a_i)} r_i,
    \label{eq:ips}
\end{equation}
where $\frac{\pi_\theta(s_i,a_i)}{\pi_0(s_i,a_i)}$ is the importance ratio, which corrects for the fact that the actions in $D$ were sampled from $\pi_0$ rather than $\pi$.
Equation \eqref{eq:ips} is known as inverse propensity score (IPS) \citep{StrehlLearning2010,Li2014Counterfactual} and although it is an unbiased estimator, it can have very high variance especially when $\pi$ deviates from $\pi_0$.
\section{Methods}\label{sec:method}
We now describe how the above techniques were adapted for wireless network optimization. We will focus on the three main challenges we described in the introduction; namely, large action space, covariate shift, and computational efficiency.

\begin{wrapfigure}{r}{0.55\textwidth}
    \centering
    \includegraphics[scale=0.35]{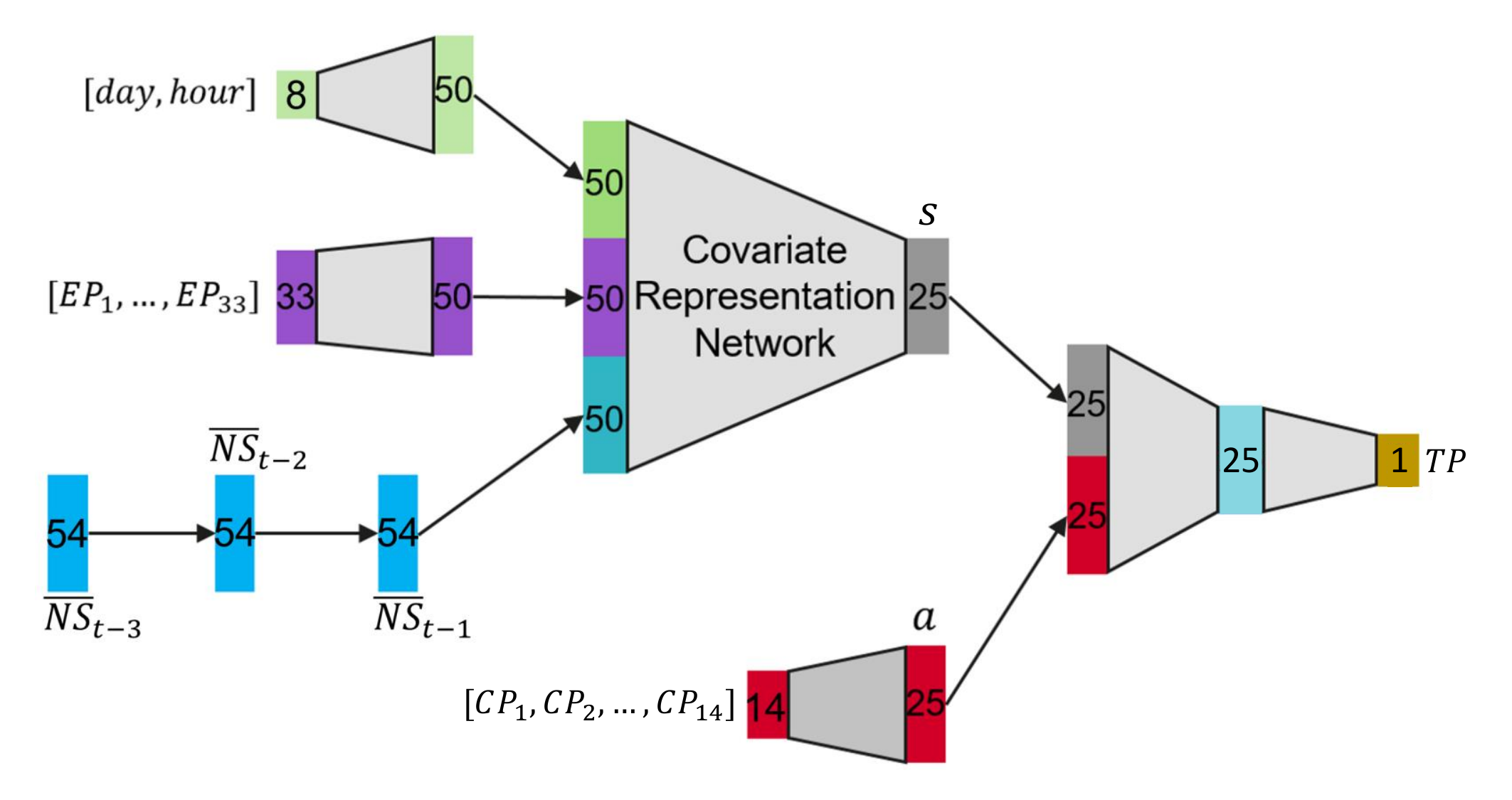}
    \caption{Reward model with modular architecture. Modularity ensures a balanced representation of all input modalities and sensitivity to CPs. Coloured rectangles indicate input covariates and representations produced by the network, where the number of neurons is specified. Trapezoids are linear layers. Trend in the counters is modelled using a Gated Recurrent Unit network (cyan blocks).}
    \label{fig:NN}
    \vspace{-15mm}
\end{wrapfigure}

We first train a neural network to predict the reward (TP) given the network state $s$ and the action $a$. The modular neural network architecture is displayed in Figure \ref{fig:NN}.  Separate input heads balance the influence of temporal context, engineering parameters (EPs) characterising a cell, and counters by mapping them to tensors of equal dimensionality (50 neurons). The ‘Covariate Representation Network’ merges the resulting tensors to yield a representation of the network state $s$. Explicitly balancing the representations of $s$ and $a$ ensures the influence of CPs is not diluted in the presence other high dimensional covariates, such as counters. Sensitivity to CPs is further enhanced by introducing them near the output layer. Finally, all representations are concatenated and passed through linear output layers to give the predicted TP. Once the network is trained, we can specify a greedy policy that selects the action $a^*$ that maximizes the predicted reward given network state $s$.
\subsection{Challenge 1: Large Action Space}\label{sec:challenge1}
The optimization problem specified in \eqref{eq:dm} can be solved using exhaustive search if the action space is finite and small. However, this is prohibitively expensive in our setting due to the high-dimensional and continuous action space. The proposed solution is to perform gradient ascent (GA) using the reward model, which is point-wise differentiable.  
We make use of software for automatic differentiation \citep{Paszke2019Pytorch} to compute $\nabla_a \hat{R}(s,a)$. Then, starting from a random initial $a_0$ we iteratively update its value following the direction of steepest ascent,
\begin{equation}
\small
    a_{k+1} = a_k + \alpha \nabla_a \hat{R}(s,a),
\end{equation}
where $\alpha \in \mathbb{R}$ constrains the steps size in the action space.
If $\hat{R}(s,a)$ is non-convex (as it is normally the case for multi-layer neural networks)  gradient ascent is only guaranteed to find a local optimum. The process is repeated multiple times with different initial values for $a_0$ to avoid poor local optima. 

\subsection{Challenge 2: Sparse Data and Covariate Shift}\label{sec:challenge2}

As mentioned in Section \ref{sec:directmethod}, since the DM is prone to bias, deploying the resulting policy directly on the real network may be unsafe.
In general, we cannot guarantee that the reward model will generalize well to regions of the state-action space that are underexplored by the logging policy \citep{Janner2019When}.
A biased reward model may confidently predict that an action will attain a high TP, even if this action was not well supported in the dataset. As a result, the policy might not necessarily select the actions for which the true reward is the largest but also those for which the model is merely highly overconfident. Two complementary solutions are introduced to alleviate bias and improve the reliability of the DM: data augmentation and uncertainty regularization. 

\subsubsection{Counterfactual Data Augmentation}
Our ability to train accurate and reliable reward models is hindered by selection bias in the factual dataset. Selection bias arises because CPs are not set uniformly at random, but rather they are generated by the logging policy, when it maps network state information to optimized CPs. The resulting dataset may lack diversity (if some CPs are never selected) and contain spurious correlations between CPs and TP (due to confounding). 


Challenges arising from selection bias motivate the proposed causal inference matching algorithm. Our method draws inspiration from \cite{schwab2018perfect, schwab2020learning}. We augment the factual dataset with hypothetical examples obtained via matching. A hypothetical example is synthesized from each factual example in the dataset. A single hypothetical example is obtained by replacing the \textit{factual} CPs (given by the logging policy) with new \textit{hypothetical} CPs sampled uniformly at random from their allowed range of values. Ten factual nearest neighbours are identified based on Euclidean distance. These nearest neighbours should have similar CPs and covariates to those of the hypothetical. The unknown TP that would have been observed if the hypothetical CPs had been executed is estimated as the mean (factual) TP of the nearest neighbours. An augmented dataset is produced in this fashion and used to train the reward model. The resulting model will generalize over arbitrary CPs, even if such CPs would not be selected by the logging policy.
\subsubsection{Uncertainty Regularization.} An uncertainty penalty is added to the objective in \eqref{eq:dm} in order to prevent the optimization process from converging to those regions of the state-action space for which the model may be biased. To do so we first create $K$ different bootstrap samples $\mathcal{D}_k$ of size $N$ by sampling with replacement from $\mathcal{D}$. A neural network ensemble \cite{Hansen1990Neural, Heskes1996Practical, Khosravi2011Comprehensive} is then trained by fitting $K$ models $\hat{R}_k(s,a)$ (with the identical architectures) to each of the $K$ datasets. Reward estimates are computed as the average prediction of all models in the ensemble. The uncertainty penalty is given by the standard deviation of those predictions,
\begin{equation}
\small
    \pi(s) = \argmax_{a \in \mathcal{A}} \left(\hat{\mu}(s,a)  - \beta \hat{\sigma}(s,a)\right)
    \label{eq:unpen}
\end{equation}
with $\mu$ being the empirical mean, $\sigma$ the standard deviation and $\beta \in \mathbb{R}$ a hyperparameter controlling the magnitude of the uncertainty penalty.
Since the objective \eqref{eq:unpen} is still differentiable, gradients can be  backpropagated through each of the models in the ensemble to compute $\nabla_a  \left (\hat{\mu}(s,a)  - \beta \hat{\sigma}(s,a) \right)$.


\subsection{Challenge 3: Computational Efficiency}\label{sec:challenge3}

One important caveat of the above described method is its computational complexity.  Running multiple iterations of GA every time an action is chosen is extremely time consuming, especially if a neural network ensemble is used to estimate the uncertainty. An alternative, is to optimize a policy network $\pi_\theta(s,a)$ directly by following the off-policy policy gradient (OPPG) \citep{Precup2000OffPolicy}. An estimate of the OPPG can be easily derived from \eqref{eq:ips},
\begin{equation}
\small
    \nabla_\theta J(\pi_\theta) \approx \frac{1}{N} \sum_{(x_i,a_i,r_i) \in \mathcal{D}} \frac{\pi_\theta(s_i,a_i)}{\pi_0(s_i,a_i)}  r_i \nabla_\theta \log \pi_\theta(s_i,a_i).
\end{equation}
Unfortunately, the importance weight $\frac{\pi_\theta(s_i,a_i)}{\pi_0(s_i,a_i)}$ has unbounded variance. Gradient estimates may grow arbitrarily large when certain state-action pairs for which $\pi_0(s,a) \approx 0$, become more likely under $\pi_\theta$.
The most straight-forward solution is to control the size of  $\frac{\pi_\theta(s_i,a_i)}{\pi_0(s_i,a_i)}$ by setting a maximum threshold \citep{Ionides2008Truncated, swaminathan2015Batch},
\begin{equation}
\small
    \nabla_\theta J(\pi_\theta) \approx \frac{1}{N} \sum_{(x_i,a_i,r_i) \in \mathcal{D}} \min \left\{ \frac{\pi_\theta(s_i,a_i)}{\pi_0(s_i,a_i)},  M\right\}   r_i \nabla_\theta \log \pi_\theta(s_i,a_i),
\end{equation}
where the hyperparameter $M$ lets us trade variance for bias. Notice that, if we set $M=1$ we get back the standard policy gradient estimator.

\subsubsection{The Hybrid Method.}
The variance problem becomes more severe when the action space is high-dimensional. This is because, as we increase the number of dimensions the likelihood of certain actions under the logging policy $\pi_0$ decreases exponentially, 
thus limiting the quality of the policies we can learn. Nonetheless, we can still benefit from these policies, even if they improve the performance of $\pi_0$ only slightly. We propose a \emph{hybrid method} whereby GA is executed on the reward model, but rather than starting from a random initial point, we start from the actions suggested by a pre-trained policy network $\pi_\theta$. Our experiments suggest that this approach not only decreases the number of random starts needed, but also helps GA converge much faster, since 
the actions sampled from $\pi_\theta$ are often good enough, and only need to be fine-tuned.

\section{Evaluation}\label{sec:evaluation}

\begin{figure}[h]
\vspace{-6mm}
\begin{subfigure}{.52\textwidth}
    \centering
    \includegraphics[width=8.5cm]{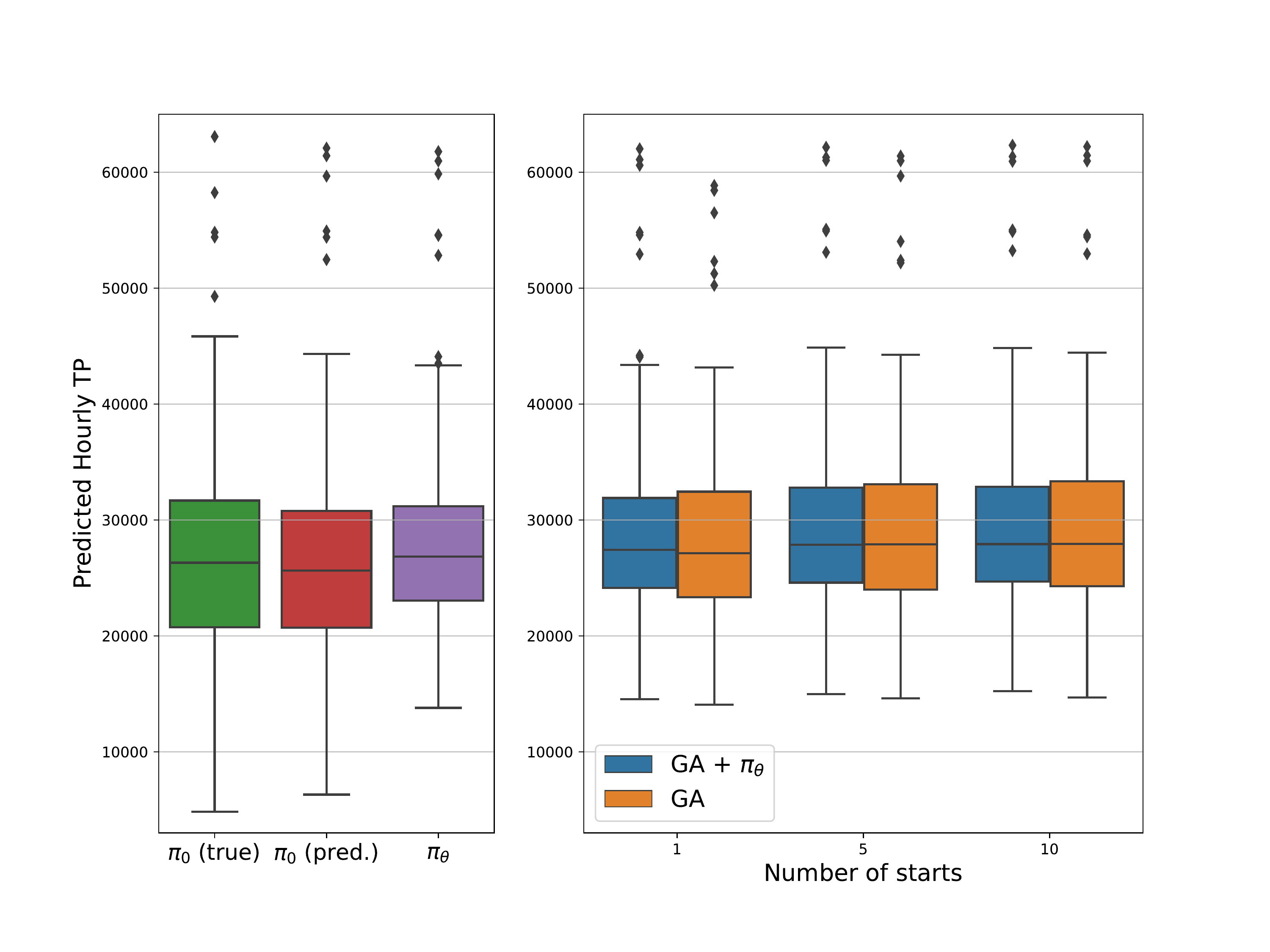}
\end{subfigure}
\begin{subfigure}{.52\textwidth}
    \centering
    \includegraphics[width=6.4cm]{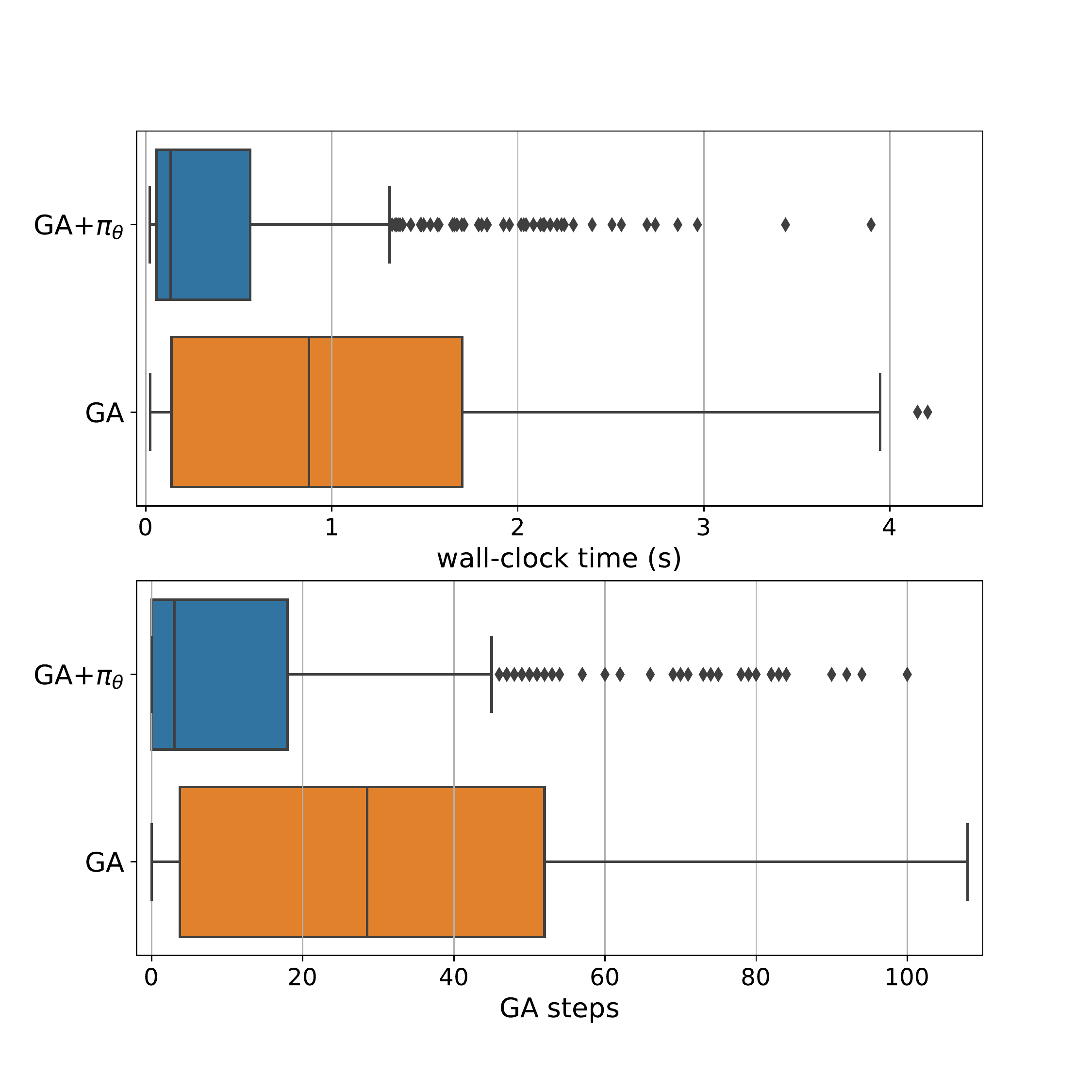}
\end{subfigure}
\vspace{-4mm}
\caption{\textbf{Left:} True (green) and predicted (red) hourly TP for the logging policy $\pi_0$. Predicted TP for the policy network $\pi_\theta$ (purple). \textbf{Middle:} Predicted TP for GA with 1, 5, and 10 starts when sampling the initial actions at random from a uniform distribution (orange) and from  $\pi_\theta$ (blue). \textbf{Top right:} Wall-clock time of running a full GA loop on the reward model with (blue) and without (orange) $\pi_\theta$. \textbf{Bottom right:} Number of GA steps to convergence with (blue) and without (orange) $\pi_\theta$.} 
\vspace{-3mm}
\label{fig:direct-method}
\end{figure}

The method is tested on 10K samples drawn from the logged data that were excluded from the training set. We follow a similar evaluation protocol to that of \cite{brookes2019Conditioning} and use two separate reward model ensembles each of them consisting of 5 neural networks trained on different bootstrap samples. We use the first ensemble $\hat{R}_\theta$ to optimize the actions and report the TP predicted by the second ensemble $\hat{R}_\theta^*$. The plot on the left of Figure \ref{fig:direct-method} shows the true TP (green) and the predicted TP given by $\hat{R}_\theta^*$ realized when CPs were set by the logging policy (red). The purple box plot shows the predicted TP when the CPs were set by the policy network $\pi_\theta$ (see Section \ref{sec:challenge3}). 

Figure \ref{fig:direct-method} (middle) compares the performance of running GA on $\hat{R}_\theta$ when choosing the initial actions at random (orange) against the hybrid method (blue), which starts from the actions suggested by the policy network (GA $+ \pi_\theta$). The box plots suggest that both GA and GA $+ \pi_\theta$ outperform the logging policy (red). As expected, the performance gain grows when we increase the number of starts. 

On the other hand, even though the policy network $\pi_\theta$ trained via OPPG as a stand-alone solution (purple) improves over $\pi_0$ (red) only slightly, when combined with GA (GA $+ \pi_\theta$) it matches the performance of standard GA with random starts while mitigating the computational burden. This is supported by the horizontal box plots on the right of Figure \ref{fig:direct-method}. The plot on the top shows the total wall clock time of running a full loop of GA on the reward model. The plot on the bottom shows the number of GA steps it takes to converge to an optimum. The results clearly indicate that the actions sampled from $\pi_\theta$ are often good enough and only need to be fine-tuned. Hence, the improvement in computational efficiency. 

\begin{figure}[h]
\vspace{-5mm}
\begin{subfigure}{.49\textwidth}
  \centering
    \includegraphics[width=7cm]{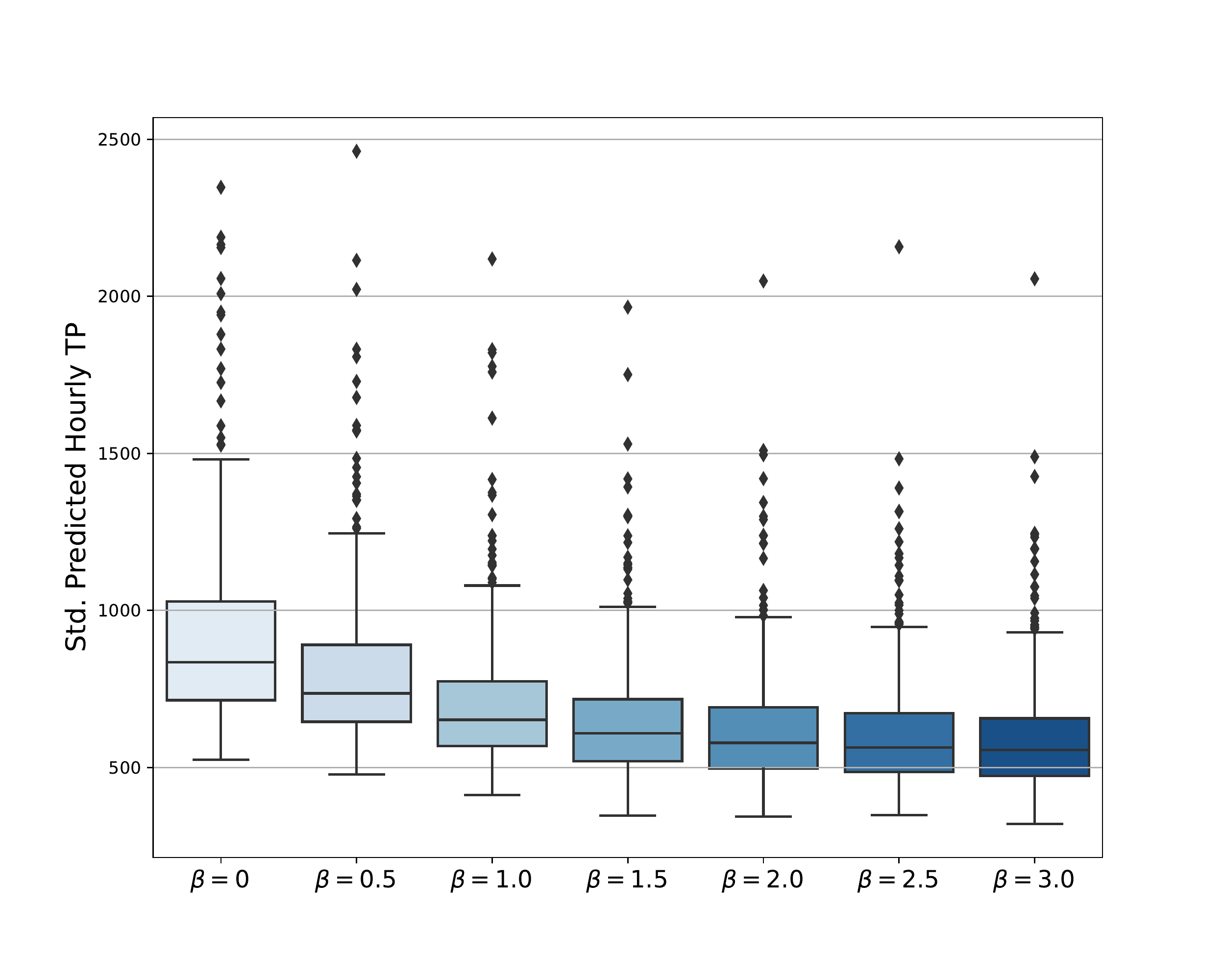}
\end{subfigure}
\begin{subfigure}{.49\textwidth}
    \includegraphics[width=7cm]{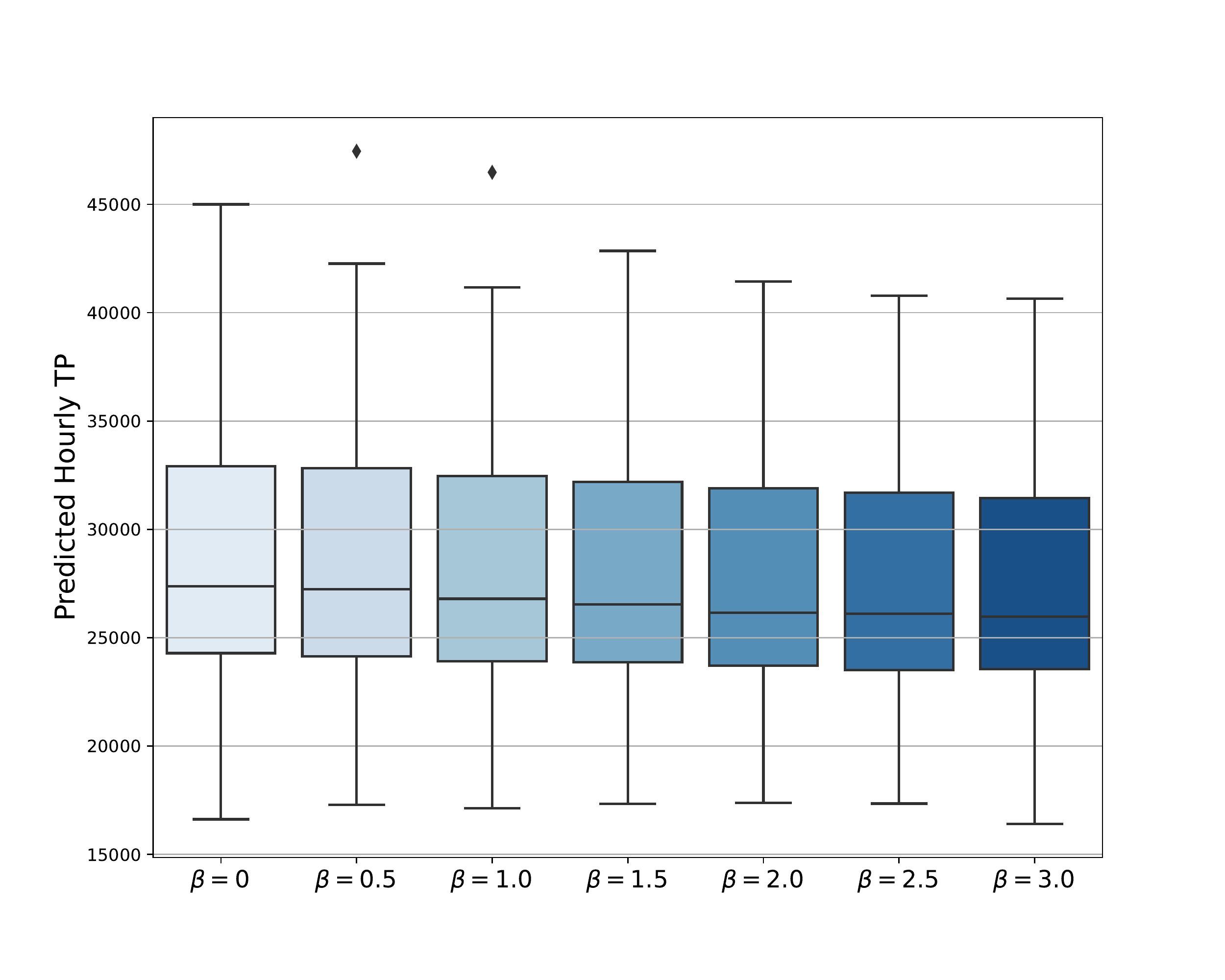}
\end{subfigure}
\vspace{-4mm}
\caption{\textbf{Left:} Uncertainty of the predicted TP after optimizing \eqref{eq:unpen} with GA for different values of penalty coefficient $\beta$. \textbf{Right:} Predicted TP after optimizing \eqref{eq:unpen} for different values of $\beta$.}
\label{fig:uncertainty-penalty}
\vspace{-5mm}
\end{figure}

The performance of GA $+\pi_\theta$ when adding the uncertainty penalty term to the objective function \eqref{eq:unpen} is reported in Figure \ref{fig:uncertainty-penalty}. The plot on the left shows how the uncertainty of the TP predictions decreases as we increase the penalty coefficient $\beta$, which implies that GA is avoiding the regions were the reward predictions might be overconfident. On the other hand, the plot on the right reveals that, decreasing the uncertainty comes at the expense of potentially lower TP gains.


\section{Conclusion}
In this paper we presented a data-driven solution to the wireless network optimization problem. We first showed how we formulated the problem as a contextual bandit, and explained how we learned policies offline from a static dataset. Our method combines a policy network trained via OPPG followed by gradient-based local fine-tuning. To guarantee robustness against model bias we augmented the dataset with hypothetical counterfactual examples. We also applied an uncertainty penalty to the optimization objective. Our results suggest that the proposed hybrid method can handle large action spaces while being computationally efficient. Moreover, pending online evaluation, 
we expect important gains in TP when this solution is deployed in the real wireless network.

\bibliography{bibliography}
\bibliographystyle{apalike}
\end{document}